\documentclass[10pt, conference, compsocconf]{IEEEtran}
\usepackage{graphicx}
\ifCLASSINFOpdf
\else
\fi
\begin{document}
%
\title{Semantic HMC for Big Data Analysis}


\author{\IEEEauthorblockN{Thomas Hassan}
\IEEEauthorblockA{Université de Bourgogne\\
Dijon, France\\
thomas.hassan@checksem.fr}
\and
\IEEEauthorblockN{Rafael Peixoto}
\IEEEauthorblockA{Polytechnic of Porto\\
Porto, Portugal\\
rafpp@isep.ipp.pt}
\and
\IEEEauthorblockN{Christophe Cruz}
\IEEEauthorblockA{Université de Bourgogne\\
Dijon, France\\
christophe.cruz@u-bourgogne.fr}
\and
\IEEEauthorblockN{\hspace{3.7cm}Aurélie Bertaux}
\IEEEauthorblockA{\hspace{3.7cm}Université de Bourgogne\\
\hspace{3.7cm}Dijon, France\\
\hspace{3.7cm}aurelie.bertaux@iut-dijon.u-bourgogne.fr}
\and
\IEEEauthorblockN{Nuno Silva}
\IEEEauthorblockA{Polytechnic of Porto\\
Porto, Portugal\\
nps@isep.ipp.pt}
}


%


\maketitle

\begin{abstract}
Analyzing Big Data can help corporations to improve their efficiency. In this work we present a new vision to derive Value from Big Data using a Semantic Hierarchical Multi-label Classification called Semantic HMC 
based in a non-supervised Ontology learning process. We also proposea Semantic HMC process, using scalable Machine-Learning techniques and Rule-based reasoning.
\end{abstract}


%
\IEEEpeerreviewmaketitle

\section{Introduction}
Nowadays, discovering knowledge and insights over web data is a major task for most corporations to increase their competitiveness. Determining the value of information relative to a particular customer is a complex task addressed by the business intelligence/data-mining field \cite{witten2005data}. In the context of Big Data, this task is even more challenging, due to its characteristics. 
An increasing number of V’s has been used to characterize Big Data \cite{Chen2014}, \cite{Hitzler2013}: Volume, Velocity, Variety and Value. Volume concerns the large amount of data that is generated and stored through the years by social media, sensor data, etc\cite{Chen2014}. Velocity concerns both to the production and the process to meet a demand because Big Data is not only a huge volume of data but it must be processed quickly. Variety relates to the various types of data composing the Big Data. These types include semi-structured and unstructured data representing 90\% of his content \cite{Syed2013} such as audio, video, webpage, and text, as well as traditional structured data, etc. Value measures how valuable the information to a Big Data consumer is. Value is the most important feature of Big Data and his “raison d’être” because if data don’t have value then is useless.
An IDC report \cite{Gantz2011} proposes the value extraction from very large volumes of a wide variety of data, by enabling the high-velocity capture, discovery, and/or analysis. Sheth \cite{Sheth2014} proposes deriving Value via harnessing the challenges posed by Volume, Variety, and Velocity using semantic techniques and technologies. This requires organized ways to harness and overcome the four V-challenges by using metadata and employ semantics and intelligent processing.
Our aim is to extract Value from Big Data by harnessing a huge Volume and Variety of data that change constantly (Velocity) by using a novel unsupervised ontology learning process based on HMC called Semantic HMC. Hierarchical Multi-Label Classification (HMC) is the combination of Multi-Label classification and Hierarchical classification \cite{Bi2011}. In HMC, the items can be assigned to different hierarchical paths and simultaneously may belong to different class labels in the same hierarchical level \cite{Bi2011}. The ontology \cite{Studer1998} plays a key role in defining terms and meanings used to represent the knowledge, reducing the gap between the users and the HMC process.
This paper does not aim to improve the state of the art in multi-classification, nor the automatic hierarchy construction. Instead it proposes a scalable process to semantically learn the ontology by adopting scalable machine learning processes and Rule-based reasoning \cite{Urbani:2011:QBR:2063016.2063063} to classify the data items and therefore extract value from Big Data.
The contributions of this work are twofold:
\begin{itemize}
\item Scalable ontology learning process based on HMC (Semantic HMC).
\item Big Data Analysis using a Semantic HMC.
\end{itemize}
The rest of the paper covers three sections. The second section describes how to use the Semantic HMC to extract value from Big Data. The third section describes the Semantic HMC process proposal. Finally, the last section draws conclusions and suggests further research.


\section{USING SEMANTIC HMC TO DERIVE VALUE IN BIG DATA CONTEXT}

\begin{figure}[!t]
\centering
\includegraphics[width=1\columnwidth]{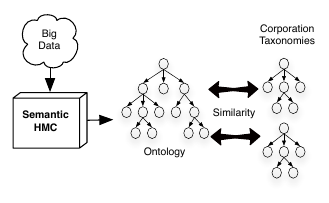}
\caption{Value extraction for corporations}
\label{fig_1}
\end{figure}

Our approach is to exploit value from very large volumes of data that are in constant generation using a Semantic HMC approach. The Semantic HMC process learns the Tbox (Taxonomy and Rules) part of the ontology from the huge Volume and Variety of initial data. Once this learning phase is finished, the classification system incrementally learns the Tbox from the new incoming items (Abox) to provide and respond to the Velocity (and the others V) dimension(s). The result of this Semantic HMC process is a rich ontology with the items (instances) classified according to the learned concept hierarchy (i.e. the taxonomy of the ontology).
Corporations recurrently use concept hierarchies as taxonomies to represent their valuable information \cite{lambe2014organising}. Our vision is to use the concept hierarchies from corporations to validate the value of the learned ontology for a specific corporation (Fig. \ref{fig_1}). Higher similarity between the concept hierarchy of the learned ontology and the concept hierarchy used by a corporation suggest better alignment between the HMC results and the corporation’s knowledge and goals. Consequently, data items classified as valuable concepts ultimately present better value to the corporation than those not matching the corporation’s concepts.

\section{SEMANTIC HMC PROCESS}

\begin{figure}[!t]
\centering
\includegraphics[width=1\columnwidth]{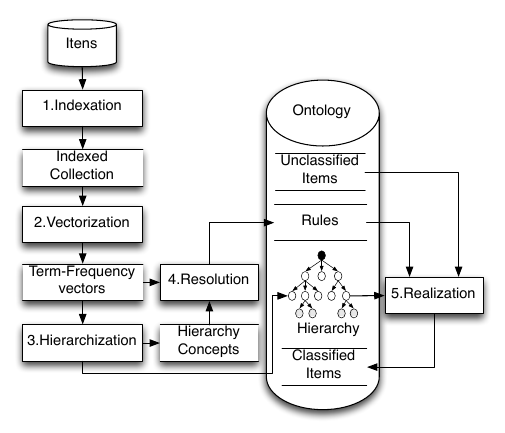}
\caption{Semantic HMC Process}
\label{fig_2}
\end{figure}

Our Semantic HMC process is generic for a large Variety of unstructured data items (e.g. text, images) and scalable for a large Volume of data. The process is unsupervised such that no previously labeled/classified examples or rules to relate the data items with the labels exist. The label (i.e. concepts) hierarchy and the rules are automatically learned from the data through scalable Machine Learning techniques. 
To infer the most specific concepts for each data item and all subsuming concepts we use rule-based reasoning that exhaustively applies a set of rules to a set of triples to infer conclusions \cite{Urbani:2011:QBR:2063016.2063063}, i.e. the items classifications. This Rule-based reasoning approach allows the parallelization and distribution of work by large clusters of inexpensive machines through Big Data technologies as Map-reduce \cite{Dean2008}. Web Scale Reasoners \cite{urbani2013three} currently uses Rule-Based reasoning to reach high scalability by load parallelization and distribution, thus addressing the Velocity and Volume dimensions of Big Data. 
This proposed process consists of 5 individually scalable steps (Fig. \ref{fig_2}) matching the requirements of Big Data processing:
\begin{itemize}
\item Indexation parses and creates an index of data items.
\item Vectorization calculates term-frequency vectors of the indexed items.
\item Hierarchization creates a concept hierarchy based on term frequency.
\item Resolution creates the reasoning rules to relate data items with the hierarchy concepts based on term frequency.
\item Realization first populates the ontology with items and then for each item determines the most specific hierarchy concept and all his subsuming concepts. 
\end{itemize}

\subsection{Indexation}
The indexation step parses and index data items. As one of our main focus points is the scalability of the architecture, the indexation is a mandatory step. Each item type has its specific parser to efficiently retrieve useful information for the other steps reducing the Limited Context Analysis problem. The Limited Content Analysis problem \cite{lops2011content}\cite{Bobadilla2013109} is defined by the difficulty in extracting reliable automated information from various content (e.g. text, images, sound, etc.), which can greatly reduce the quality of the classifications. By reducing the Limited Content Analysis we improve the Semantic HMC capability to handle more Variety of data.

\subsection{Vectorization}
The vectorization step vectorizes the terms in the indexed items by calculating two types of term frequency vectors :
\begin{itemize}
\item Term frequency in each item using the frequency of a term in an item measured by TF-IDF. TF-IDF uses the frequency of a term in an item (TF) and the inverse number of items in which the term appears (IDF) \cite{Salton1988}.
\item Term frequency in all items using the appearing frequency of a term in all documents \cite{Salton1988}.
\end{itemize}
The following steps use the term vectors calculated in this step to learn the concepts Hierarchy and the Rules. 

\subsection{Hierarchization}

The hierarchization step will select relevant terms as relevant concepts and also will generate the broader-narrower relations between these concepts. To select the concepts in the hierarchy, a quality measure must be used. 
There are several methods for creating hierarchical relations between concepts including \cite{DeKnijff2013}, \cite{Meijer2014}:
\begin{itemize}
\item Hierarchical clustering that starts with one cluster and progressively merges clusters that are closest. 
\item Subsumption methods that construct the concept broader-narrower relations based on the co-occurrence of concepts.
\end{itemize}
Any of these methods can be used to create the hierarchical relations. The advantages and drawbacks of each method is deeply studied in \cite{DeKnijff2013}.

\subsection{Resolution}

The resolution process will create the ontology rules used to relate the hierarchy concepts and the items using the term-frequency vectors. The rules creation process will use thresholds as proposed in \cite{Werner2014} to select the most relevant terms for each hierarchy concept that will be used in the rules. The main difference is that instead of translating the rules into logical constraints of an ontology captured in Description Logic, these rules will be translated into rules in the Semantic Web Rule Language (SWRL). The main interest in using SWRL rules is to reduce the reasoning effort, thus improving the scalability and performance of the system. We aim to use a huge amount of simple SWRL rules that will be applied to the ontology in order to classify items. 

\subsection{Realization}

The realization phase will populate the learned concept hierarchy with data items. First the ontology is populated with new items to label in an assertion level (Abox). To do the classification/labeling of the items, the SWRL Rules generated in the Resolution step are used. Then a Rule-Based inference engine will use the SWRL rules and the hierarchy to infer the most specific concepts for each data item and all subsuming concepts. This leads to a multi-label classification of the documents based in a hierarchy of labels (Hierarchical Multi-label Classification).

\section{Conclusion}
In this paper we present our vision to extract value from Big Data using a Semantic HMC process and propose a scalable five-step architecture to automatically classify unstructured items. We use machine learning to learn an ontology with SWRL rules to automatically classify items of Big Data. The Semantic HMC process prototype is under development and we expect to show the implementation and results in further work.
Our current work consists in evaluating the resulting ontology, considering three different aspects: the process scalability (performance), the quality of the hierarchy, and the quality of the classification process (i.e. concept tagging of items).


\section*{Acknowledgment}

This project is founded by the company Actualis SARL, the French agency ANRT and through the Portuguese COMPETE Program under the project AAL4ALL (QREN13852).


\newcommand{\BIBdecl}{\setlength{\itemsep}{0em}}
\bibliographystyle{IEEEtran}
\bibliography{IEEEBigData2014}

%



\end{document}